\documentclass[journal]{IEEEtran}
\ifCLASSINFOpdf
  \usepackage[pdftex]{graphicx}
  \graphicspath{{../pdf/}{../jpeg/}{./images/}}
  \DeclareGraphicsExtensions{.pdf,.jpeg,.png}
\else
  \usepackage[dvips]{graphicx}
  \graphicspath{{./eps/}}
  \DeclareGraphicsExtensions{.eps}
\fi

\usepackage{amsmath}

\usepackage{xcolor}
\usepackage[justification=centering]{caption}

\usepackage{algorithm}
\usepackage{algorithmic}

\usepackage{array}

\usepackage{tabularx}

\begin{document}
%
\title{Weak Human Preference Supervision \\ For Deep Reinforcement Learning\\}
%
%
%

\author{Zehong~Cao,~\IEEEmembership{Member,~IEEE,}
        KaiChiu~Wong,~\IEEEmembership{}
         Chin-Teng~Lin,~\IEEEmembership{Fellow,~IEEE}

\thanks{Z. Cao is with the School of Information and Communications Technology (ICT), University of Tasmania, Australia. (E-mail: zehong.cao@utas.edu.au.)}%
\thanks{K. Wong was with the University of Tasmania and is now with the MyState Bank, Australia.}%
\thanks{C.T. Lin is with the Australian Artificial Intelligence Institute (AAII) and the School of Computer Science, University of Technology Sydney, Australia.}

\thanks{Manuscript received on 27 July, 2020; revised on 06 December, 2020.}}

\markboth{IEEE Transactions on Neural Networks and Learning Systems}%
{Shell \MakeLowercase{\textit{et al.}}: Bare Demo of IEEEtran.cls for IEEE Journals}
%

\maketitle

\begin{abstract}

The current reward learning from human preferences could be used to resolve complex reinforcement learning (RL) tasks without access to a reward function by defining a single fixed preference between pairs of trajectory segments. However, the judgement of preferences between trajectories is not dynamic and still requires human input over thousands of iterations. In this study, we proposed a weak human preference supervision framework, for which we developed a human preference scaling model that naturally reflects the human perception of the degree of weak choices between trajectories and established a human-demonstration estimator via supervised learning to generate the predicted preferences for reducing the number of human inputs. The proposed weak human preference supervision framework can effectively solve complex RL tasks and achieve higher cumulative rewards in simulated robot locomotion -- MuJoCo games -- relative to the single fixed human preferences. Furthermore, our established human-demonstration estimator requires human feedback only for less than 0.01\% of the agent's interactions with the environment and significantly reduces the cost of human inputs by up to 30\% compared with the existing approaches. To present the flexibility of our approach, we released a video (https://youtu.be/jQPe1OILT0M) showing comparisons of the behaviours of agents trained on different types of human input. We believe that our naturally inspired human preferences with weakly supervised learning are beneficial for precise reward learning and can be applied to state-of-the-art RL systems, such as human-autonomy teaming systems.

\end{abstract}

\renewcommand\IEEEkeywordsname{Keywords}


\IEEEpeerreviewmaketitle

\section{Introduction}
Reinforcement learning (RL) \cite{mnih2015human} typically uses a reward function to train an agent's behaviours for a specified task. Nevertheless, constructing an effective reward function in complicated scenarios can often be challenging. If the design of a reward function is too simple, then the behaviours of the trained agent may not match our expectations, i.e., the results may exhibit misalignment between our expectations and the actual testing \cite{nikolaidis2017game}. To achieve more effective RL, having a communication pathway between the RL agent and our expectations during the training process is valuable \cite{bogert2016expectation}. 

In autonomous vehicle research, robotic controls with RL have already demonstrated tremendous potential in improving transportation systems for related problems such as localisation, path planning, and collision avoidance \cite{lu2020bearing} \cite{xu2019design}. Investigating human-autonomy teaming using RL can enable the development of a reliable reward function in the human-robot interaction process. Some recent work showed challenges in training an intelligent robot to complete task objectives \cite{russell2016should} \cite{amodei2016concrete} \cite{ke2020enhancing} and in multi-agent interactions \cite{cao2019reinforcement} \cite{cao2020hierarchical}, addressing the difficulties of alignment between human expectations and the final training outcome. Although approaches such as inverse RL \cite{RN55} and imitation learning \cite{RN59} were suggested to extract the reward function and mimic the actions of human experts to ensure an expected outcome, these approaches are not direct enough to train the desired behaviour. Moreover, the degree of movement of a robot could be larger than that of a human, as a human demonstration for imitation learning may not be available in some cases \cite{schroecker2017state}.

\begin{figure}[!t]
\centering
\includegraphics[width=3.4in]{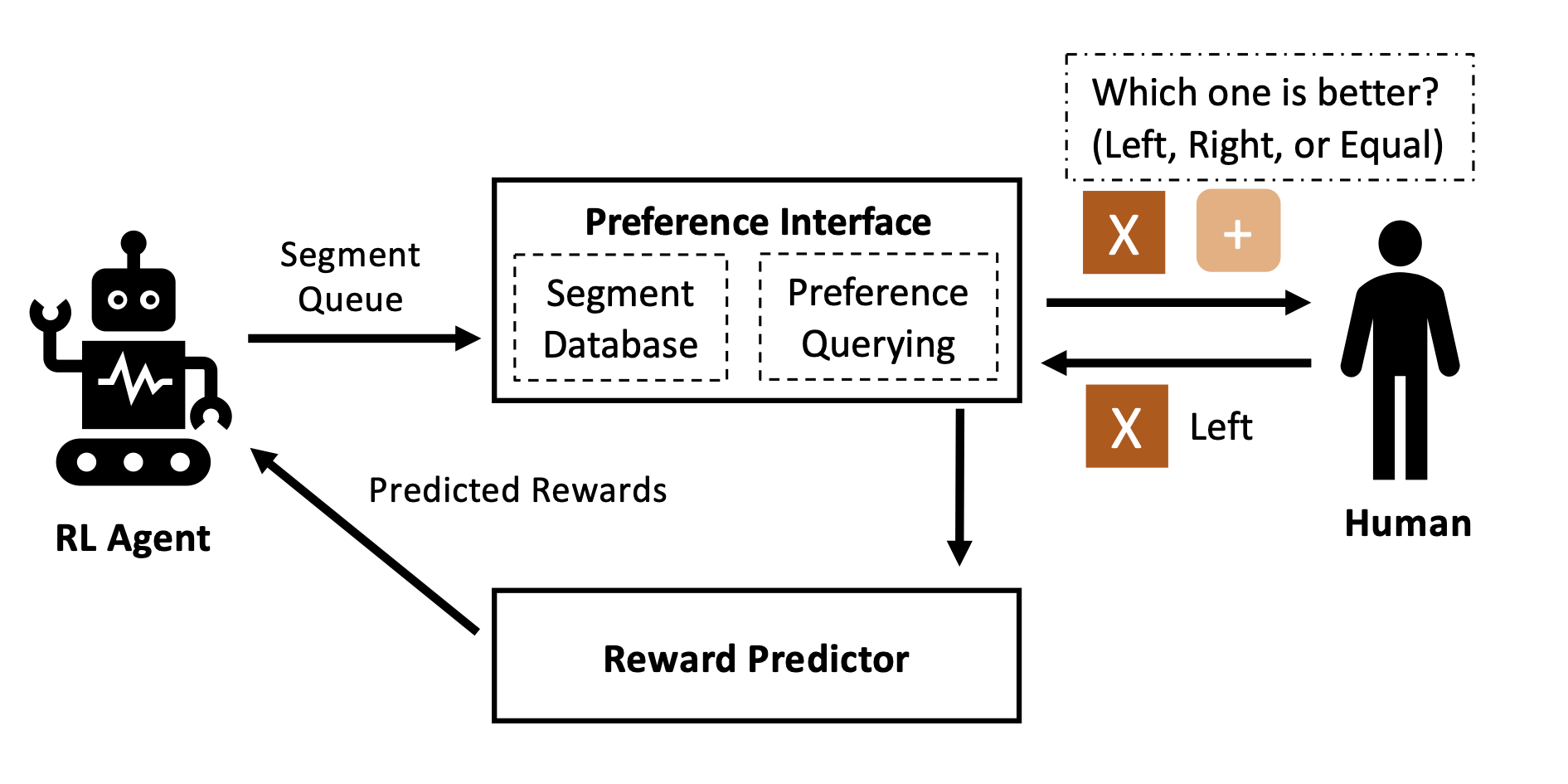}
\caption{Illustration of Deep RL from Human Preferences}
\label{hprl}
\end{figure}

A novel study proposed by the DeepMind group \cite{RN52} first addressed deep RL from human preferences, in which a comparison measurement between pairs of trajectory segments was designed to replace the reward function and preference inputs from an expert were allowed, as shown in Figure \ref{hprl}. This approach requests advice from an expert to ensure that the RL training is on the correct track, implying that agents could be assisted when tackling problems in some highly complicated scenarios. However, the judgement of preferences between trajectories is not dynamic. For example, the candidate preferences include only fixed left, right, or equal options, represented by the preference values 1, 0, and 0.5, respectively; thus, we assume that this approach cannot reflect natural human intentions. In addition, the current approach still requires human input over thousands of iterations, which requires a substantially large amount of time from humans.

Inspired by the above DeepMind study and the uncertainties in decision making \cite{RN52} \cite{xiao2020novel}, in this study, we propose a new framework: weak human preference supervision for deep RL. In particular, we develop a scaling model to support dynamic and weak human preferences, instead of single fixed preferences, for RL. Moreover, we use a database of human preference scaling values to establish a human-demonstration estimator via supervised learning to predict the preference scaling values based on initial human inputs to reduce the amount of human effort. 

The contributions of our weak human preference supervision framework for RL are as follows:

\begin{enumerate}
\item Our proposed weak human preference supervision for deep RL allows humans to input dynamic and weak preference levels via our developed human preference scaling model to reflect human behaviour and decrease the number of human inputs for our established human-demonstration estimator.

\item Based on 5 experiments with the robotic physical simulator MuJoCo \cite{RN61}, our developed human preference scaling model for RL can achieve higher cumulative reward values than those of the current fixed human preference model, and our established human-demonstration estimator support for RL can reduce the amount of human input for dynamic and weak preferences by up to 30\% without significantly sacrificing the reward values.  

\end{enumerate}

The rest of this paper is organised as follows. The related work is briefly introduced in Section II. Then, the preliminaries and our proposed framework are illustrated in Section III. Afterwards, experiments involving MuJoCo games and the relevant settings are addressed in Section IV. Finally, we present our findings and comparison results in Section V and conclude this work in Section VI.

\section{Related Work}

\subsection{Preference Learning}
An agent may often not be able to learn expected actions from rewards when using the traditional RL strategy, but preference learning can potentially minimise the gap between the missing agent information and the behaviour desired by humans. A recent study \cite{woodworth2018preference} stated that preference learning could facilitate the adaptation of robotic movement and noted that the human demonstration is particularly easy for task-dependent goals. An earlier study \cite{furnkranz2012preference} also showed that human preferences are more effective in acquiring better actions than are agent rewards in RL. They implied that the feedback from a human could leverage RL in qualitative policy models and that manipulation from preference learning could be a new strategy to assist agents in achieving the desired human behaviour in robotic control. 

Additionally, \cite{aggarwal2019modelling} provided a human behaviour decision model to increase the accuracy in preference learning, but it focused only on adjusting the attitude and the importance of relative criteria without improving the human preference coverage. Another recent work \cite{palan2019learning} developed a DemPref model to provide efficient preference-based learning in robotic movements, but this approach requires collecting a large number of purely human-demonstrated actions prior to training. 

\subsection{Learning by Pairwise Comparisons}
Learning by pairwise comparisons allows the application of human preferences to RL. F{\"  u}rnkranz's work \cite{furnkranz2012preference} revealed that the preferences could help an agent carry out label ranking in RL so that humans could provide feedback to the agent. To distinguish between preference learning and RL, a survey study \cite{wirth2017survey} clarified that an agent can receive feedback from two options in preference learning, while in RL, only one human feedback option is accepted. This restriction facilitates RL agent label ranking when performing a task. 

$z_i \succ z_j$ indicates that $z_i$ is the preferred choice over $z_j$, where $ \mathit{Z} $ represents a set of options. Kreps \cite{kreps1988notes} offered some notations that could be defined from preference learning while comparing a pair of options:

\begin{itemize}
\item $z_i \succ z_j$: option $z_i$ is \textit{absolutely preferred}.
\item $z_i \prec z_j$: option $z_j$ is \textit{absolutely preferred}.
\item $z_i \sim z_j$: options $z_i$ and $z_j$ are \textit{the same}, as no preferred option can be distinguished. 
\item $z_i \succeq z_j$: option $z_i$ is \textit{weakly preferred}. 
\item $z_i \preceq z_j$: option $z_j$ is \textit{weakly preferred}. 
\end{itemize}

\subsection{RL from Human Preferences}
To link preference learning to RL, according to the communication pathway proposed by Christiano \cite{RN52}, the basic flow of RL from human preferences comprises three main modules: agent, preference interface, and reward predictor. The agent continually trains and explores the environment as RL progresses. The novelty of the communication pathway is the addition of a preference interface, which randomly generates some episodes in which human judgement is requested. An expert therefore inputs which options he/she prefers by inputting the preference database into the preference interface. These preferences are sent to a reward predictor to perform further training to assess the relative rewards from these preferences. These data are then used return the trained preferences as the agent’s observations to perform the policy training.

\subsection{Motivations}
The settings of Christiano's study \cite{RN52} cover only three conditions among the definitions of preference learning from Kreps \cite{kreps1988notes}, i.e., $z_i \succ z_j$, $z_i \prec z_j$ and $z_i \sim z_j$. Christiano's study did not consider $z_i \succeq z_j$ and $z_i \preceq z_j$, which are two weak but important conditions we encounter in real-life decision making based on preferences. This lack motivates us to develop a new RL-based human preference setup to cover the weak preference conditions $z_i \succeq z_j$ and $z_i \preceq z_j$.

Furthermore, Christiano's study \cite{RN52} required a large number of human preference labels, which may require a long period of time dedicated to human inputs. In particular, at least 5,500 human inputs are required in the training setup. We assume that some human preference labels may not be accurate, as a human may experience a high level of fatigue, potentially resulting in an increase in the error rate after a long period of performing the preference judgement task. This problem motivates us to develop a human-demonstration estimator via supervised learning to reduce the number of human inputs and improve learning performance.

\section{Preliminaries and Methods}

\subsection{Settings and Goal}
An RL agent interacts with the environment for a number of steps, during which the agent inspects the environment according to the observation $o_{t} \in O$, receives an instant reward $r$ at each timestep $t$ and then performs action $a_{t} \in A$ based on the observation. The agent aims to maximise the cumulative rewards during the training and receives the predicted instant reward values during each timestep $r_{t} \in R$ via the reward predictor and preference interface modules. When the agent takes an action in the environment, a video clip from a trajectory segment is pushed into the list in the preference interface. The preference interface module randomly draws two video clips from the list and asks a human which clip he/she would prefer. Once the human inputs his/her preference, this information is passed to the reward predictor to generate predicted rewards for the agent for training. 

In this study, we follow the preference interface design from Christiano's work \cite{RN52}, which accepts the generated segments from the RL agent and places them into a queue, as shown in Figure 1. Two segments are randomly selected from the queue, and the preference interface asks a human for his/her preference between the two candidate options.
After the preference interface collects the preferred option from the human, this option is saved in the preference queue of the reward predictor for the later training of RL policies. Because the currently used approach accepts only a fixed preference, i.e., left, right or equal, as its input, it lacks dynamic inputs of the preferred range. In this study, we modify the preference interface design to consider weak human preference conditions and propose a synthetic preference scaling model, as shown in the following section.

\subsection{Weak Human Preferences: Synthetic Preference Scaling}

In this study, we assume that the human always chooses the trajectory segment that has the most potential to return a high reward value; thus, we use the synthetic oracle, a Bayesian approach for policy learning from trajectory preference queries, to mimic the preference of the human, whose preference over several trajectories precisely reflects the reward \cite{wilson2012bayesian}. For the synthetic human preference, when the agent queries for comparisons, the synthetic human can immediately reply by indicating a preference for whichever trajectory segment yields a higher reward in the underlying task. 

Based on the synthetic human preference \cite{wilson2012bayesian}, we develop a weak human preference model called synthetic preference scaling, as presented in Algorithm \ref{algor1}.

\begin{algorithm}
\caption{Weak Human Preferences: Synthetic Preference Scaling}
\label{algor1}
\begin{algorithmic}[1]

\REQUIRE synthetic preference scaling represented as $\widehat{z}$.
\REQUIRE $\widehat{z}$ from a given reward set ($R_{left}$, $R_{Right}$) in the range [0.0, 1.0], where equal preference corresponds to a value of 0.5.

\renewcommand{\algorithmicrequire}{\textbf{Input:}}
\renewcommand{\algorithmicensure}{\textbf{Output:}}
\REQUIRE {$R_{left}$}, the reward of the left trajectory segment.
\REQUIRE {$R_{right}$}, the reward of the right trajectory segment.
\REQUIRE {$R$}, the reward list composed of all $n$ reward sets.

\ENSURE  {$\widehat{z}$}
\STATE $ R \gets \left [ R_{left_{1}}, R_{right_{1}}, R_{left_{1}}, R_{right_{1}}, ... ,  R_{left_{n}}, R_{right_{n}}  \right ] $
\STATE sort($R$), sort the list, in ascending order by default.
\STATE $N \gets $  number of elements in $R$.
\STATE $R_{min}  \gets  \left ( \left \lceil \frac{10}{100}\times N \right \rceil \right )^{th} $ element from $R$.
\STATE $R_{max}  \gets  \left ( \left \lceil \frac{90}{100}\times N \right \rceil \right )^{th} $ element from $R$.
\IF {($R_{left} > R_{right}$)}
\STATE normalise $\widehat{R}_{left}  \gets  max\left ( 0,min\left ( \frac{R_{left} - R_{min}}{R_{max} - R_{min}},1 \right ) \right ) $
\STATE $\widehat{z}  \gets  0.5 + 0.5 \times \widehat{R}_{left} $
\ELSIF{($R_{left} < R_{right}$)}
\STATE normalise $\widehat{R}_{right}  \gets  max\left ( 0,min\left ( \frac{R_{right} - R_{min}}{R_{max} - R_{min}},1 \right ) \right ) $
\STATE $\widehat{z}  \gets  0.5 - 0.5 \times \widehat{R}_{right} $
\ELSE
\STATE $\widehat{z} \gets 0.5$
\ENDIF
\RETURN $\widehat{z}$
\end{algorithmic}
\end{algorithm}

We elaborate on the proposed Algorithm \ref{algor1} in the following. Because the range of ${z}$ is set as [0.0, 1.0], a ${z}$ value of 1.0 means that the human absolutely prefers the left trajectory segment, while a ${z}$ value of 0.0 means that the human absolutely prefers the right trajectory segment. A ${z}$ value of 0.5 indicates that the human cannot judge between the two trajectory segments. For preference scaling, the reward list $R$ is collected in the memory in each iteration, and the synthetic preferences are calculated based on the reward values with a normalisation measurement. In particular, based on the 90\% confidence interval level, from the sets in the reward list $R$, we first remove the lowest 10\% and highest 10\% of reward values to ensure that outliers or anomalous reward values will not affect the preference scaling calculation. Then, all collected reward values are normalised to values between 0.0 and 1.0, and all sets in the reward list $R$ are ranked in ascending order. 

\subsection{Human Preference Scaling for Deep RL}

Following the basic preference settings for RL as addressed in Section III-A, at each time $t$, we maintain the policy $\pi$ : $\mathit{O} \rightarrow \mathit{A}$, where the agent interacts with the environment according to observation $o_t \in \mathit{O} $ and then performs particular action $a_t \in \mathit{A}$ based on instant observation $o_t$. During the training, the agent tries to estimate the reward function $\widehat{r}$ : $\mathit{O} \times \mathit{A} \rightarrow \mathit{R}$ from a deep neural network, which is updated as follows:

\vspace*{1\baselineskip} 

\begin{enumerate}
\renewcommand{\labelenumi}{(\theenumi)}

\item A set of trajectories $\left \{ \gamma^{1}, \gamma^{2}, ... , \gamma^{i} \right \}$ is generated by policy $\pi$. The parameters of policy $\pi$ are updated by the traditional RL to ensure that the maximum sum of predicted rewards $r_{t} = \widehat{r}\left ( o_{t}, a_{t} \right )$ that could be achieved from observation $o$ and action $a$ is obtained. 

\item A pair of segments $\left ( \sigma^{1}, \sigma^{2} \right )$ is randomly selected from a set of trajectories $\left \{ \gamma^{1}, \gamma^{2}, ... , \gamma^{i} \right \}$. This pair of segments is sent to the preference interface and allows the human to perform the comparison.

\item Based on Algorithm \ref{algor1}, the preference scaling $z$ is collected from the human and linked to the pair of segments $\left ( \sigma^{1}, \sigma^{2} \right )$.
\end{enumerate}

Please note that the above updating processes occur in the asynchronous mode: process (1) passes the trajectories $\left \{ \gamma^{1}, \gamma^{2}, ... , \gamma^{i} \right \}$ to process (2), process (2) passes the human preferences to process (3), and process (3) passes the parameters of $\widehat{r}$ back to process (1).

\begin{figure*}[!t]
\centering
\includegraphics[width=7.4in]{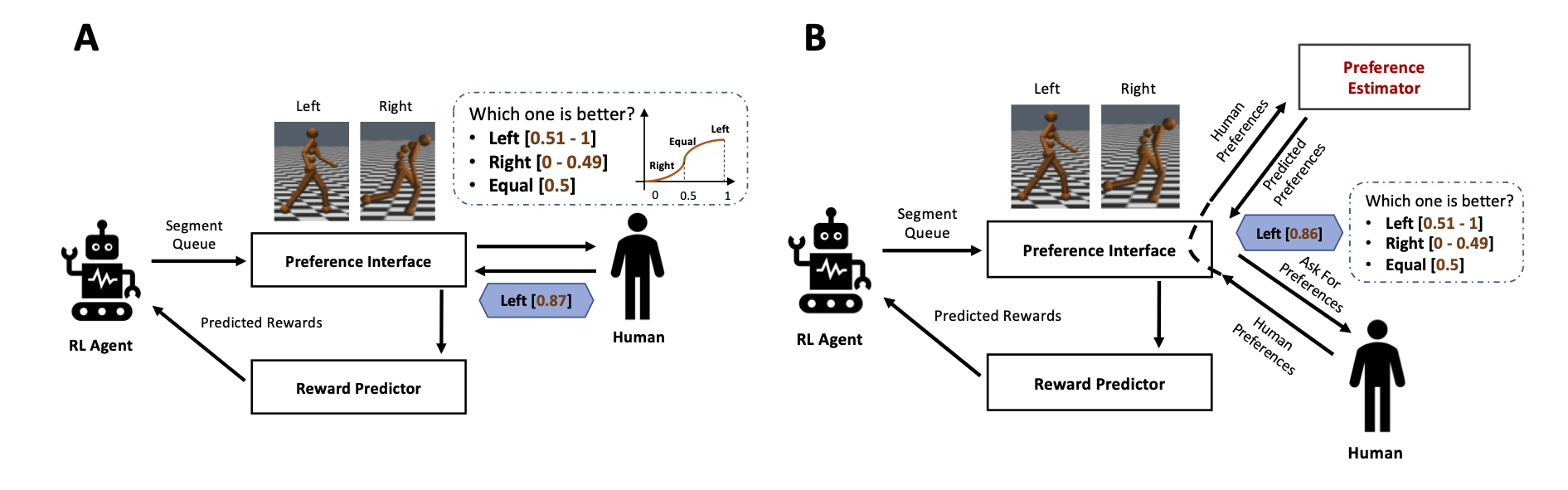}
\caption{Proposed (A) Human Preference Scaling Model and (B) Human-Demonstration Model for Deep RL}
\label{shprl_harl}
\end{figure*}

\vspace*{1\baselineskip} 

\subsubsection*{Preference Elicitation}

In Figure \ref{shprl_harl}-A, we show a human preference scaling structure for the RL agent to make these processes easily understandable. To reflect natural human intent, we modify the format of human preferences from fixed preferences to scale-based preferences in process (3). In terms of the current fixed preferences for a pair of segments \cite{RN52} (as shown in Figure \ref{hprl}), this approach allows inputting only the left, right or equal option, and the judgement is saved in the preference database $D$ with the data format $ (\sigma^{1}, \sigma^{2}, {z})$, where $\sigma^{1}$ and $\sigma^{2}$ are the extracted paired segments and ${z} = [{0, 0.5, 1}]$ is the fixed preference input from the human. Thus, the approach does not provide information regarding how much a human prefers a particular segment. For example, the left segment could be better than the right segment, but the left one may not be perfect. 

Our proposed scale-based preferences provide a scaling model (as shown in Algorithm \ref{algor1}) with which the human can input a dynamic score for the preferred segment by assigning any value between 0.0 and 1.0. The value of $\widehat{z}$ could be in the range of $\left \{ 0.0, 1.0 \right \}$ to specify the dynamic judgement of the human. If $\widehat{z}$ is input as 0.0, then the condition of $\sigma^{1}$ is absolutely preferred, while if $\widehat{z}$ is input as 1.0, then the condition of $\sigma^{2}$ is absolutely preferred. A $\widehat{z}$ value of 0.5 indicates that $\sigma^{1}$ and $\sigma^{2}$ are not different; hence, the above conditions do not hold. Then, we can further address the additional conditions, which are in the weakly preferred categories. To be more specific, any values between 0.0 and 0.5 (excluding the margin values 0.0 and 0.5) indicate that $\sigma^{1}$ is weakly preferred, while any values between 0.5 and 1.0 (excluding the margin values 0.5 and 1.0) indicate that $\sigma^{2}$ is weakly preferred. For example, a human could input a preference value of 0.87 or any value he/she likes in the range between 0.0 and 1.0 to specify that they weakly prefer the left or right segment. Our contribution aims to supply more accurate information regarding the human preferences to the RL agent, not simply indicating whether the chosen option is better than the other but rather indicating how much better it is. 

\vspace*{1\baselineskip} 

\subsubsection*{Fitting the Reward Function}

If the reward function estimate $\widehat{r}$ is the predicted reward from the reward predictor, as shown in Figure \ref{shprl_harl}-A, then we consider $\widehat{r}$ as a latent factor explaining the human judgements and assume that the human probability of preferring a segment $\sigma$ depends exponentially on the value of the latent reward summed over the length of the clip, which follows Christiano's design process \cite{RN52}, as follows.
$$\hat{P}\left[\sigma^{1} \succ \sigma^{2}\right]=\frac{\exp \sum \hat{r}\left(o_{t}^{1}, a_{t}^{1}\right)}{\exp \sum \hat{r}\left(o_{t}^{1}, a_{t}^{1}\right)+\exp \sum \hat{r}\left(o_{t}^{2}, a_{t}^{2}\right)}$$
$\widehat{r}$ minimises the cross-entropy loss between these predictions and the actual labels of human inputs:
$$\operatorname{loss}(\hat{r})=$$
$$-\sum_{\left(\sigma^{1}, \sigma^{2}, \widehat{z}\right) \in \mathcal{D}} \widehat{z}(1) \log \hat{P}\left[\sigma^{1} \succ \sigma^{2}\right]+\widehat{z}(2) \log \hat{P}\left[\sigma^{2} \succ \sigma^{1}\right]$$

\vspace*{1\baselineskip} 

\subsubsection*{Optimising the Policy}

After the reward function $\widehat{r}$ computes the rewards, we can meet the need for traditional RL. As the reward function, $\widehat{r}$ may be non-stationary, which leads us to prefer RL algorithms that are robust to changes in the reward function, such as policy gradient methods \cite{ho2016model}. In this study, we use proximal policy optimisation (PPO) \cite{schulman2017proximal} to perform simulated robotics tasks and apply the same parameter settings as in Christiano's work \cite{RN52}.

\subsection{Human Preference Supervision: Human-Demonstration Estimator for Deep RL}

Because a large number of human preference scaling values (generally over 1,000 preference inputs) must be stored in the preference database to train the fitting of predicted rewards, we assume that reducing the number of human preference inputs by training a preference estimator is worthwhile. As shown in Figure \ref{shprl_harl}-B, we develop a preference estimator based on the previous human inputs and use a regression model with supervised learning that we call the human-demonstration estimator to predict some of the human preferences. We expect this human-demonstration estimator to not only reduce the number of human inputs $n$ but also maintain good performance without sacrificing the cumulative rewards.

The human-demonstration estimator is an extended version of the previous human preference scaling for deep RL presented in Section III-C. In particular, the collected database includes $n$-fold preference scalings of a preference estimator based on previous human inputs, and a regression model with supervised learning that we call the human-demonstration estimator is used to predict some of the human preferences with the data format $ (\sigma^{1}, \sigma^{2}, \widehat{z})$, as discussed earlier. Specifically, the initial preferences are collected at the initial stage and stored in the preference queue. These preferences are used as the base of the preference estimation. In our study, two reliable supervised learning models, i.e., linear regression and support vector regression (SVR) with a radial basis function (RBF) kernel, are considered as the estimator to predict the human preference based on the collected initial preferences. To construct a prediction model and fit the parameters of the human preference scaling estimator, the database is separated into two parts, i.e., a training dataset and a testing dataset, and we implement two types of data splitting: i) 50\% training data and 50\% testing data and ii) 70\% training data and 30\% testing data. As a result, 30-50\% of the human inputs will be replaced by the agent's estimation. 

With linear regression, the objective function for ordinary least squares with one preferred segment $\sigma$ in the set $ (\sigma^{1}, \sigma^{2})$ is as follows:

$$\operatorname{Min} \sum_{i=1}^{n}\left(\widehat{z}'_{i}-w_{i} \sigma_{i}\right)^{2}$$
where $\widehat{z}'_{i}$ is the estimated preference scaling value and $w_{i}$ is the coefficient.

SVR gives us the flexibility to define errors and finds an appropriate hyperplane in higher dimensions to fit the data. The objective function of SVR is to minimise the coefficients: the l2-norm of the coefficient vector. The error term is instead handled in the constraints, where we set the absolute error less than or equal to a specified margin, called the maximum error $\varepsilon$. We can tune $\varepsilon$ to gain the desired accuracy of our model. The updated objective function of SVR and the constraints are as follows:

Minimise:
$$
\operatorname{Min} \frac{1}{2}\|\boldsymbol{w}\|^{2}
$$

Constraints:

$$
\left|\widehat{z}'_{i}-w_{i} \sigma_{i}\right| \leq \varepsilon
$$
where $\widehat{z}'_{i}$ is the estimated preference scaling value and $w_{i} $ and $\sigma_{i}$ are the coefficients and the preferred segment, respectively.

To evaluate the performance of the estimators, the mean squared error (MSE) is applied to measure the average of the squares of the errors, i.e., the average squared difference between the estimated preference scaling values $\widehat{z}' $ and the ground truth of preference scaling $\widehat{z}$. 

\section{Experiment}

We implemented the existing models and our proposed models for deep RL and performed experiments in 5 scenarios from MuJoCo \cite{todorov2012MuJoCo} with TensorFlow \cite{abadi2016tensorflow} under the OpenAI Gym platform \cite{brockman2016openai}. The collected results were consolidated under the TensorBoard package from TensorFlow.

\subsection{Robotic Control Scenarios}

OpenAI Gym provides baseline environments to train the agent on the RL algorithms \cite{brockman2016openai}. MuJoCo is one of the popular continuous control tasks in OpenAI Gym, with a physics engine that can be used to simulate model-based control \cite{RN61}. MuJoCo \cite{todorov2012MuJoCo} contains diverse scenarios with robot control, where the agent moves different joints with continuous control instead of intermittent control to achieve a particular goal \cite{RN64}. The agent performs different types of actions to achieve the maximum cumulative reward value to reach the target goal. This process could be challenging, as the MuJoCo environment involves high exploration dimensions for the agent.

\begin{table}[]
\caption{List of 5 MuJoCo Scenarios}
\label{tab:MuJoCo_scenario}
\begin{tabular}{|p{3.7cm}|l|p{2.5cm}|}
\hline

\textbf{MuJoCo Scenario} & \textbf{Observation} & \textbf{Task Summary}\\ \hline

\textbf{Walker} \includegraphics[width=1.5in]{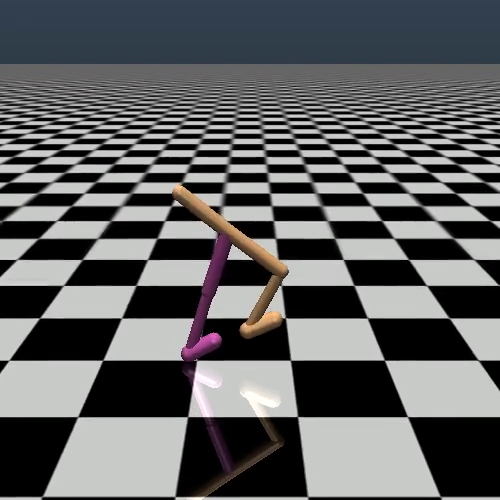} & (18, 6, 24) & A planar walker tries to roll forward and walk as quickly as possible. The reward depends on the velocity $v$ and the torso height $h$. \\ \hline

\textbf{Hopper} \includegraphics[width=1.5in]{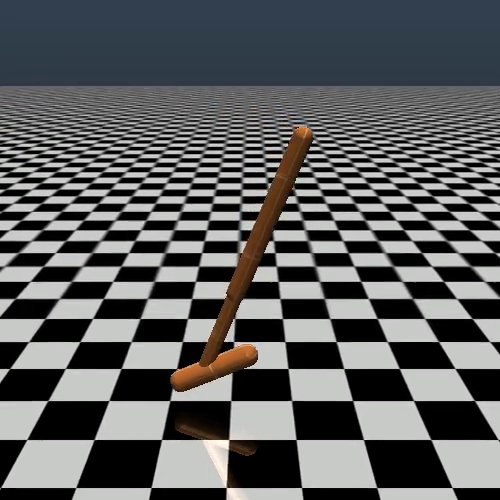}  & (14, 4, 15) &  A one-legged robot is required to move forward and attain as high a torso height as possible. The reward depends on the velocity $v$ and the torso height $h$. \\ \hline

\textbf{Swimmer} \includegraphics[width=1.5in]{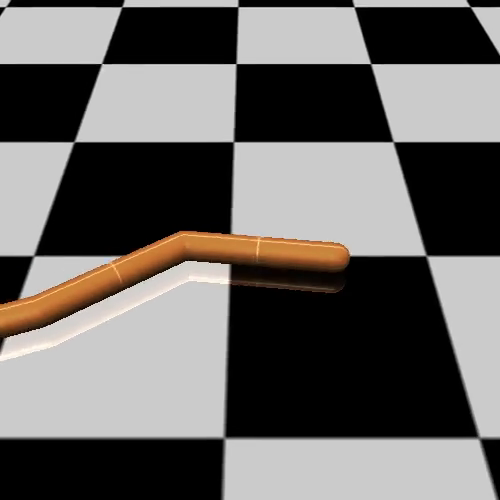} & (10, 2, 13) & A robot tries to reach a random target by swimming. The reward is given when the nose of the robot touches the random target.  \\ \hline

\textbf{Ant} \includegraphics[width=1.5in]{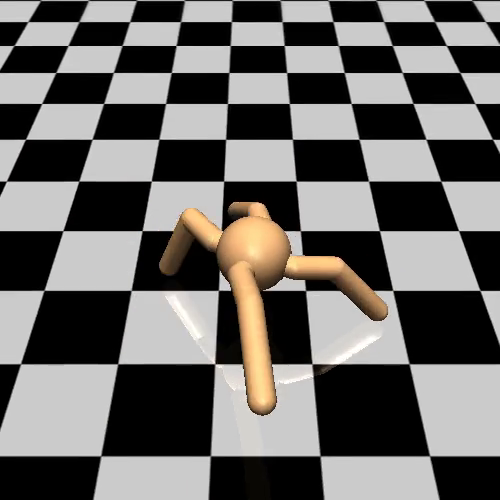} & (29, 8, 67) & A 4-legged robot attempts to learn to walk as quickly as possible. The reward is based on the velocity $v$ and the body height $h$. \\ \hline

\textbf{Cheetah} \includegraphics[width=1.5in]{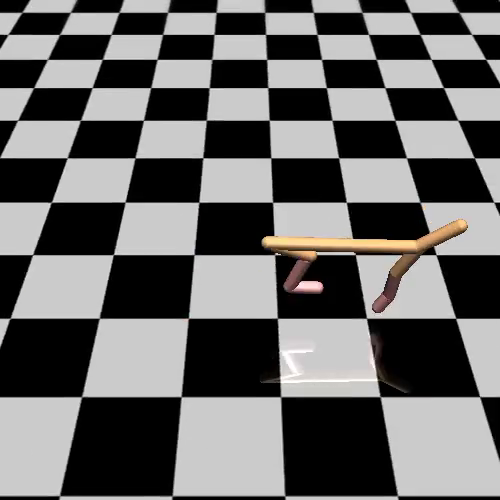} & (18, 6, 17) & A robot has to learn to move forward as quickly as possible. The reward $R$ is based on the velocity $v$, the formula for which is
$R(v)=max\left ( 0, min\left ( \frac{v}{10}, 1 \right ) \right )$.\\ \hline

\end{tabular}
\end{table}

As shown in Table \ref{tab:MuJoCo_scenario}, in this study, our testing environments include 5 scenarios: Walker, Hopper, Swimmer, Cheetah, and Ant. The existing approaches, such as traditional RL (PPO) and RL from human preferences (RLHP), and our proposed weak human preference supervision framework composed of two models, i.e., 1) weak human preferences: RL from human preference scaling (RLHPS) and 2) human preference supervision: RL from human preference scaling with demonstrations (RLHPS with Demo), are applied to compare their performances in these 5 scenarios.

\subsection{Parameter Settings}

Generally, in the policy gradient strategy, e.g., PPO, the agent starts with the initial policy, interacts with the environment, obtains a predicted reward from human feedback or from using the pre-defined reward function, and then uses the reward to improve the policy. Here, we need to know how many transitions (sequences of states, rewards, and actions) the agent should gather before updating the policy and how to use the transitions for updating under the new policy. 

First, we need to address the collection of experience (horizon, mini-batches, and epochs) prior to updating the policy. For example, PPO collects trajectories up to the time horizon ($T$) limit and then performs a minimum batch size stochastic gradient descent (SGD) update on all collected trajectories within a specified epoch. Second, to update the new policy from the old policy, PPO uses a surrogate loss function to keep the step from the old policy to the new policy within a safe range; here, consideration of the discount factor $\gamma$ and the GAE parameter $\lambda$ is necessary. In addition, the remaining parameters are general hyperparameters that can be used in many deep learning experiments to determine, e.g., the learning rate, number of steps, and number of hidden units. In this study, the experiments in the MuJoCo scenarios are trained for $3 \times 10\textsuperscript{7}$ timesteps over 10 iterations. All MuJoCo scenarios are trained under the PPO strategy with or without human feedback with the hyperparameters given in Table \ref{tab:MuJoCo_param}.  

\begin{table}[]
\centering
\caption{Hyperparameters of PPO used in MuJoCo Scenarios}
\label{tab:MuJoCo_param}
\begin{tabular}{|p{3cm}|p{3cm}|}
\hline

\textbf{Hyperparameter} & \textbf{Value} \\ \hline
Horizon ($T$) & 2048 \\ 
Mini-batch size & 64  \\ 
Number of epochs & 10  \\ 
$\gamma$ & 0.99  \\ 
GAE parameter ($\lambda$)  & 0.95  \\ 
Adam step size & $ 3 \times 10\textsuperscript{-4} $ \\ 
Learning rate & $1 \times 10\textsuperscript{-4}$ \\
Number of steps & $3 \times 10\textsuperscript{7}$ \\ 
Number of hidden units & 64 \\ \hline
\end{tabular}
\end{table}

\subsection{Baselines}

\vspace*{1\baselineskip} 
\subsubsection*{Traditional RL}

The baseline of each scenario is training with the traditional RL (PPO) without any human involvement. The agent has to learn, based on the scenario goal, only from the rewards they receive. The learning performance is the same as that of the traditional RL process and relies on the design of the reward function of each scenario. The details of the reward design of each scenario are specified in Table \ref{tab:MuJoCo_scenario}. Our goal for this setup is to set the baseline algorithm and evaluate the performance in each scenario to check what reward values could be achieved without human input.

\vspace*{1\baselineskip} 
\subsubsection*{RL from Human Preferences (RLHP)}

We consider RLHP as another baseline that contains the basic human preferences to replicate the results under advice from a human \cite{RN52}. The experimental setup emphasises a preference interface to ask for human preferences, and the user interface shows the rewards and the video clip information to let each human input a large number of preferences. The interface allows the user to input either a left, right or equal option into the preference interface.

\section{Results}

In this study, 5 scenarios (Walker, Hopper, Swimmer, Cheetah and Ant) are used as our experimental environments to evaluate the training performance among 4 types of RL model by comparing the traditional RL baselines (without human preferences), i.e., PPO and RLHP, with our two proposed models, i.e., RLHPS and RLHPS with Demo. The RL algorithms involved in this study are summarised in Table \ref{tab:RL_Algorithms}, where human preferences require an input of 700 to 1,400 labels, which correspond to less than 0.01\% of the training timesteps. 

\begin{table*}[]
\caption{Summary of the RL Algorithms}
\label{tab:RL_Algorithms}
\begin{tabular}{l|l|l|l|l}
\hline
\textbf{RL Type}                   & \textbf{Algorithm}    & \textbf{\begin{tabular}[c]{@{}l@{}}No. of Inputs \\ (1)\end{tabular}} & \textbf{\begin{tabular}[c]{@{}l@{}}No. of Inputs \\ (2)\end{tabular}} & \textbf{Note}                                                                                                                                                                              \\ \hline
RL without Human Inputs            & Traditional RL (PPO) & N/A                                                                   & N/A                                                                   &                                                                                                                                                                                            \\ \hline
                                   & RLHP                  & 1,400                                                                 & 700                                                                   & RL from human preferences.                                                                                                                                                                  \\ \cline{2-5} 
RL with Human Inputs (Preferences) & \textbf{RLHPS (ours)}           & 1,400                                                                 & 700                                                                   & RL from human preference scaling.                                                                                                                                                         \\ \cline{2-5} 
                                   & \textbf{RLHPS with Demo (ours)} & 980                                                                   & 700                                                                   & \begin{tabular}[c]{@{}l@{}}RL from human preference scaling \\ with demonstrations; \\ 30-50\% of the 1,400 inputs (420-700 inputs)\\ are generated from the estimated preferences.\end{tabular} \\ \hline
\end{tabular}
\end{table*}

\subsection{RL from Human Preference Scaling (RLHPS)}

\begin{figure*}[!t]
\centering

\begin{tabular}{l}
\includegraphics[width=7.2in]{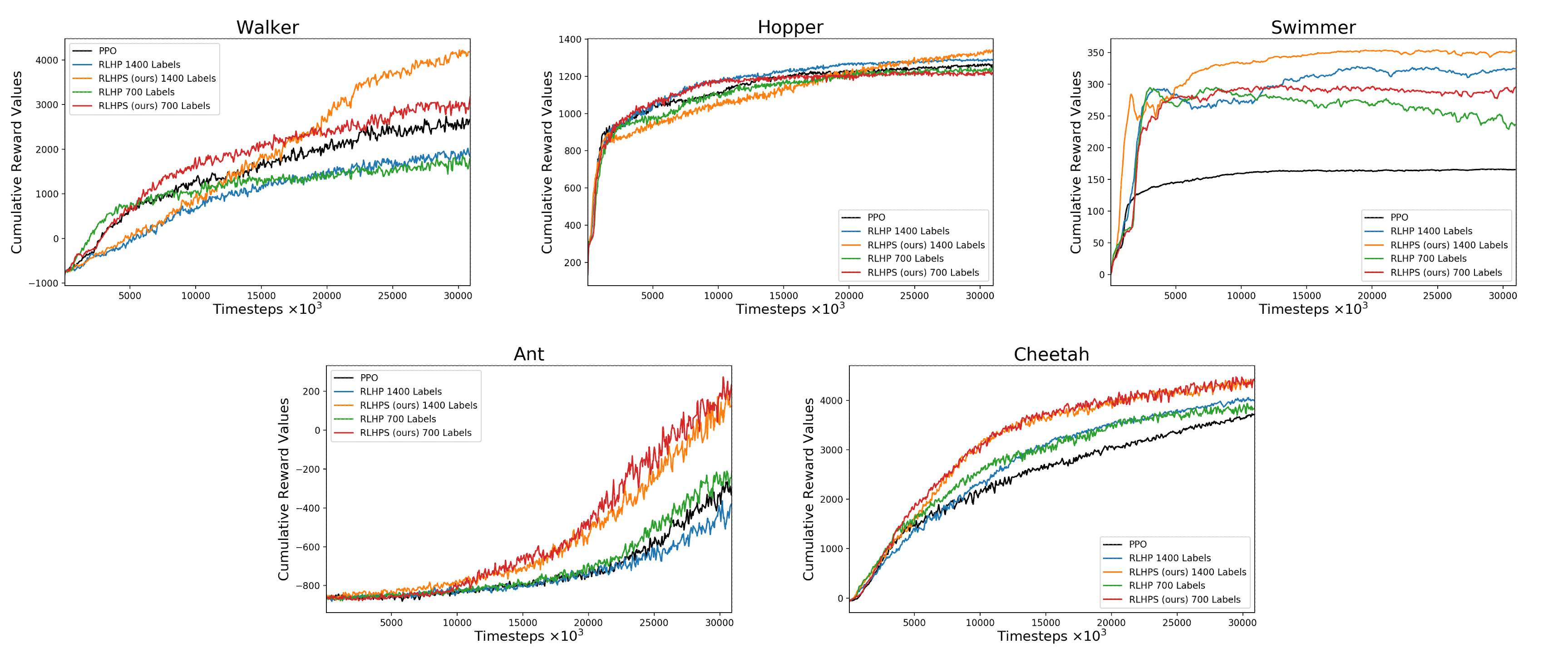}
\end{tabular}

\caption{Experimental Results of PPO, RLHP, and RLHPS in the 5 MuJoCo Scenarios}
\label{shprl_result} 
\end{figure*}

Based on the experiments in the 5 MuJoCo scenarios, the performance in terms of the cumulative reward values of the two baselines, i.e., the traditional RL (PPO) and RLHP, and our proposed approach (RLHPS) are compared, as shown in Figure \ref{shprl_result}, which presents the training of our agent by learning from two types of human preference input (700 and 1,400 labels, amounting to less than 0.01\% of the training timesteps) for RLHP and our proposed RLHPS. The cumulative reward values from our proposed RLHPS are always higher than those from RLHP or PPO, except in the Hopper scenario, where all the RL algorithms achieve similar reward values after $2 \times 10\textsuperscript{7}$ timesteps. From another perspective, the use of 1,400 human input labels generally yields higher rewards than obtained from 700 human input labels, suggesting that more human effort may benefit the training of a robust RL agent.

\vspace*{1\baselineskip} 
\subsubsection*{Cumulative Reward Values}

Particularly in the Walker scenario, our proposed RLHPS can achieve an approximately 1,500 higher reward value than the traditional RL PPO setup and an approximately 1,000 higher reward value than RLHP for the case of 1,400 human preference inputs. For the Swimmer scenario, the reward learning from RLHPS with the 1,400-label setup can achieve an approximately 350 reward value at the end of the experiment, while RLHP with the 1,400-label setup can achieve a reward value of only approximately 300 at the end of the experiment. Both human preference setups (RLHP and RLHPS) are much better than the traditional RL setup (PPO), as PPO achieves a reward value of only 150. In terms of the Cheetah scenario, our proposed RLHPS can also acquire higher rewards, a reward value of approximately 4,000, compared to the range of 3,000-3,500 when trained by RLHP and PPO.

Regarding the special case of the Ant scenario, complete three-dimensional movement and observation are required for the robot to achieve learning. Continuous movement of the agent is very difficult to achieve, as the ant robot has to find a way to balance its body and walk. From our observation, during the initial period of training, the robot does not know how to walk and balance and always flips over, which causes the reward values in this period to always be negative. The experimental results show that the scale-based preferences, highlighted in red, could yield a good performance compared to the fixed preferences. This outcome is good evidence that our proposed RLHPS can attain higher rewards (either with 700 or 1,400 labels) and quickly achieve positive reward values compared with RLHP or PPO.

\vspace*{1\baselineskip} 
\subsubsection*{Instant Reward Distributions}

For our proposed RLHPS, we also investigated the instant reward distributions between the initial and final training periods in the 5 MuJoCo scenarios. The initial training period is that in which the human has not input a preference label, generally within the first 300 timesteps. The final training period includes the final 2,500-3,000 timesteps, during which the human inputs 1,400 preference labels and the RL agent is approaching completion of the training process. This scenario was considered because the experimental setup requires use of the initial stage (the initial training period) to generate segments for the segment queue for the preference interface so that the agent can perform the initial RL training before the preference interface can select some segments from which the human can choose. We also intend to replicate the performance from Christiano’s work \cite{RN52} as our baseline; thus, our proposed method uses exactly the same settings to contrast with the existing results. 

\begin{figure*}[!t]
\centering

\begin{tabular}{l}
\includegraphics[width=7.5in]{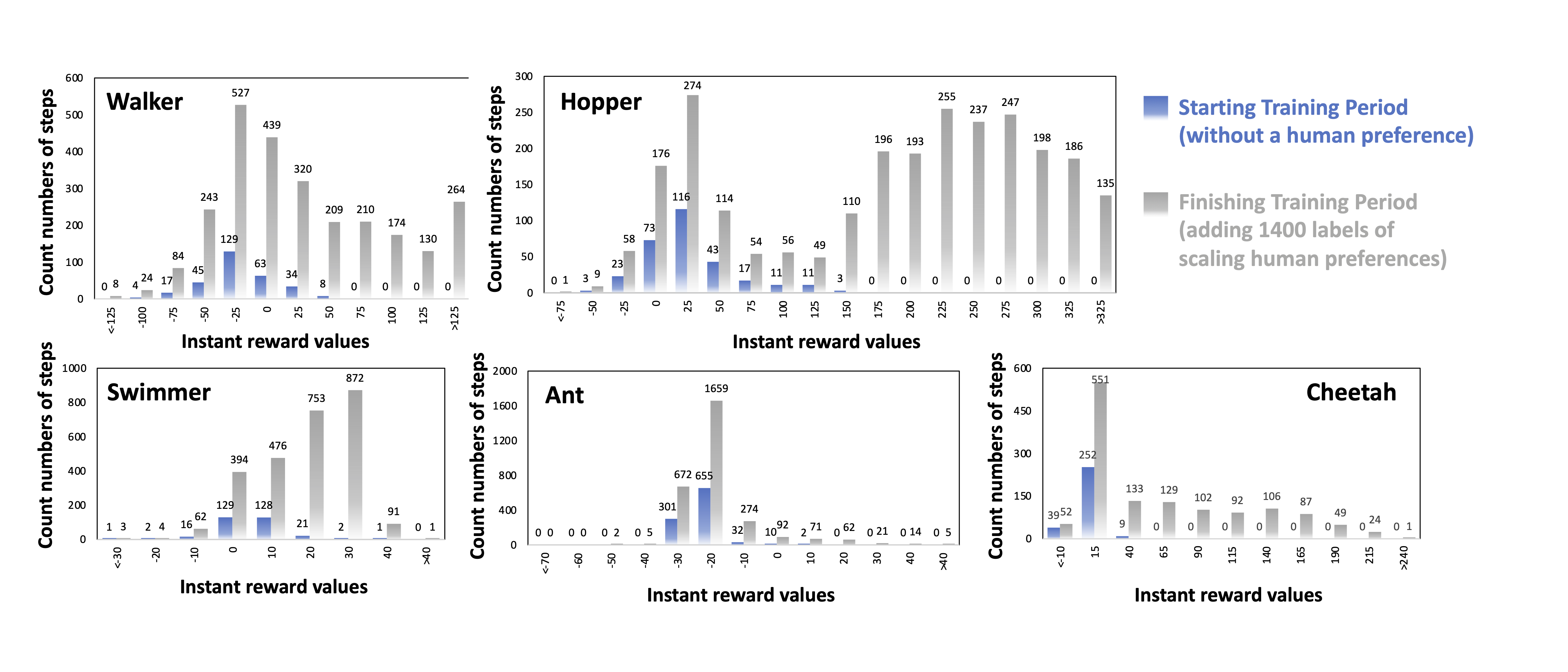}
\end{tabular}

\caption{RLHPS: Instant Reward Distributions Between the Initial and Final Training Periods in the 5 MuJoCo Scenarios}
\label{reward_result}
\end{figure*}

As shown in Figure \ref{reward_result}, the instant reward distributions are sampled for these two training periods (initial vs. final) in the 5 MuJoCo scenarios. The blue data distribution represents the beginning of training before acquiring any preferences; hence, most of the instant rewards are still negative. The grey data distribution shows the situation at the end of training, where the RL agent can achieve more positive rewards from human preferences. Our findings from the instant reward distributions confirm that our proposed RLHPS has a positive effect on RL and that the agent can use self-management to learn the values of the human preference scaling.

In summary, our findings show that the setting with human preference scaling is always much better than the baseline with fixed human preferences, which confirms that our scaling model enables the agent to learn in higher-dimensional environments, as our setting did reflect natural human intent. Additionally, from the comparison results of the instant reward distributions, RLHPS is confirmed to have excellent training performance, as instant reward values are made more favourable than those in the initial stages prior to the agent receiving any preferences. 

\subsection{RL from Human Preference Scaling with Demonstrations (RLHPS with Demo)}

\begin{figure*}[!t]
\centering
\begin{tabular}{l}
\includegraphics[width=7.2in]{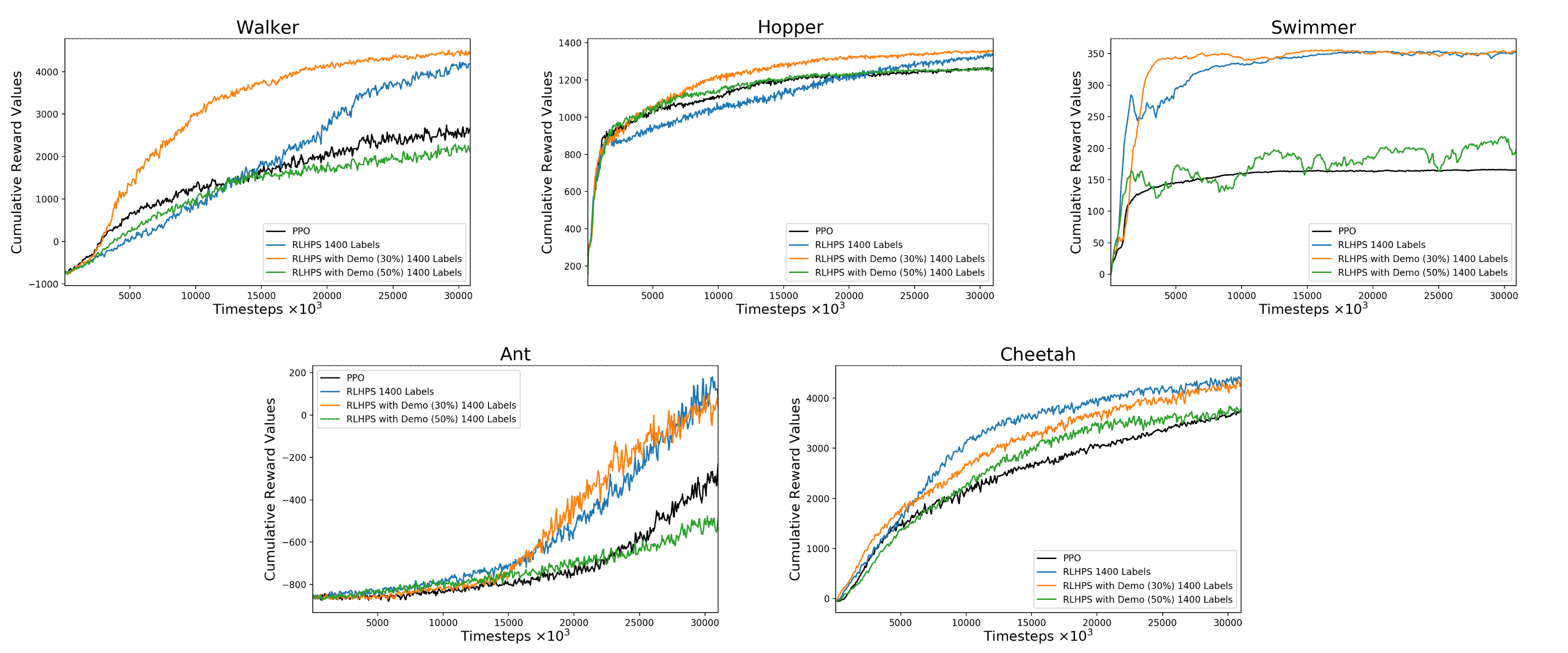}
\end{tabular}
\caption{Experimental Results of PPO, RLHPS, and RLHPS with Demo in the 5 MuJoCo Scenarios}
\label{harl_result}
\end{figure*}

After the experiments on RLHPS, we implemented the estimator interface to link the preference interface and the human based on our proposed approach -- RLHPS with Demo. As addressed in Section III-D, we performed two types of data splitting, with 30\% and 50\% of the data from the preference database used for testing. Because 1,400 human preference inputs could achieve better performance across the different scenarios according to the previous experiment, we kept the same number of labels -- 1,400 -- to test the performance of RLHPS with Demo. As shown in Figure \ref{harl_result}, the experiment employing RLHPS with Demo (30\%) indicates that we can reduce the human preference inputs by 30\%; thus, humans need to input only 70\% of the 1,400 (980) preference labels, with the remaining 30\% predicted by the regression model. The experiment employing RLHPS with Demo (50\%) aims to provide further observations of the influence of reducing the number of human preferences on the cumulative reward values, indicating that 50\% of the 1,400 (700) preference labels need to be input by the human, while 50\% of the 1,400 preferences can be predicted by linear regression or the SVR model with the smallest MSE, as shown in Table \ref{tab:mse_results}, which presents the prediction accuracies achieved using the averaged MSE trained by linear regression or SVR (with the RBF kernel) in the 5 scenarios.

\vspace*{1\baselineskip} 
\subsubsection*{Cumulative Reward Values}
Figure \ref{harl_result} generally indicates that RLHPS with Demo (30\% estimated human preferences) can achieve similar or superior cumulative reward values compared to the other approaches, i.e., RLHPS with Demo (50\% estimated human preferences) and RLHPS (without Demo), which excludes the human-demonstration stage. For RLHPS with Demo (50\% estimated human preferences), where we reduce the number of human preference inputs by half, only a performance similar to that of PPO can be achieved, far from the achievement when using RLHPS. The main reason is that removing 50\% of the human inputs will cause higher MSEs, which may lead to a negative impact on training performance.

In terms of the Walker and Hopper scenarios, RLHPS with Demo (30\% estimated human preferences) yields the highest cumulative reward values. For the Swimmer and Ant scenarios, RLHPS with and RLHPS without Demo have similar performances, suggesting that we can employ 30\% less human inputs to achieve similar outcomes. Only in the Cheetah scenario was RLHPS with Demo unable to achieve a comparable performance to that of RLHPS, suggesting that the estimated preferences may not be of benefit to walking guidance for the Cheetah scenario. 

\begin{table}[!t]
\centering
\caption{MSEs of Preference Estimations}
\label{tab:mse_results}

\setlength{\tabcolsep}{4mm}{
\begin{tabular}{l|cc|cc}
\hline
 &\multicolumn{2}{c|}{\textbf{Linear Regression}}&\multicolumn{2}{c}{\textbf{SVR (RBF kernel)}}\\

\textbf{Scenario} & Mean & Standard & Mean & Standard \\ 
 & & Deviation & & Deviation \\ 
\hline
Walker&0.0514&0.0752&0.0099&0.0184\\
Hopper&0.0688&0.0383&0.0122&0.0716\\
Swimmer&0.0239&0.0397&0.0737&0.0665\\
Ant&0.0356&0.0846&0.0637&0.1210\\
Cheetah&0.0607&0.0769&0.0725&0.0870\\
\hline
\end{tabular}}

\end{table}

Although the 30\% reduction may not show a significant improvement in terms of the percentage, it could reduce human input by a notable amount (from 1,400 to 980 inputs) without impacting the training performance. In short, it removes the need for approximately 420 human inputs per experiment. Assuming that a human needs 1 second per input on average, this reduction will amount to 7 minutes saved per experiment under the MuJoCo scenarios. A total of 25 minutes per experiment could also be saved under the Atari scenarios. If we repeat the experiments 25 times, we could save 175 minutes and 625 minutes for the MuJoCo and Atari scenarios, respectively, which are significant time reductions when considering complex scenarios.

\subsection{Observation of Testing Behaviours}

We released a video (https://youtu.be/jQPe1OILT0M) to demonstrate the behaviours of the trained agent in all MuJoCo scenarios. Generally, our proposed method, RLHPS or RLHPS with Demo can perform more appropriate behaviours to achieve the expected goal than can RLHP or PPO. In only a few exceptional cases, the goal of the Walker scenario is for the agent to roll forward and walk. Nevertheless, by observing the behaviour of the trained agent, we find that the agent operated very slowly to achieve this goal. In the Ant scenario, our RLHPS with Demo seemed to take more time to learn to walk around than required by RLHP or PPO.

\subsection{Limitations}

This study has some limitations. Our findings provide insights into some human preference labels that could be generated by the prediction model, which could reduce the number of human inputs without sacrificing the training performance while maintaining an excellent reward return. However, this reduction process is still limited. From the experimental results, when half of the human preference labels are predicted, the training performance is dramatically impacted. Therefore, under the current settings, using up to 30\% of the estimated preference inputs in place of human inputs can prevent this negative impact and maintain the training performance.

Furthermore, the frame selections are made on a random basis, which may influence the human preferences for some cases close to the equal option. We suggest that this could be refined by having an intelligent selection model select segments that have diverse rewards or at time points when the prediction model is unable to judge between the two segments. In addition, the normalisation step of our human preference scaling model could be further investigated since the distribution of rewards could affect preference levels for the estimator.

\section{Conclusions}

Our study proposed a weak human preference supervision framework involving a human preference scaling model with demonstrations that aims to effectively solve complex RL tasks and achieve higher cumulative rewards in simulated robot locomotion -- MuJoCo games. We attempted to optimise RLHP in two ways: enhancement of the weak human preference details by scaling the preference levels and reduction of the number of human preference inputs by replacing some inputs with preference labels generated by an estimator. Remarkably, our two developed models, RLHPS and RLHPS with Demo, achieve higher cumulative reward values and significantly reduce the cost of human inputs up to 30\% compared to PPO and RLHP. To present the flexibility of our approach, the released video shows comparisons of the behaviours of agents trained on different types of human input. 

Given the high scalability of deep RL, we believe that our proposed weak human preference supervision framework, which includes RLHPS and RLHPS with Demo, can help an agent learn natural human preferences with fewer inputs to enhance the training performance. In this work, our contribution to the improvement of RL-based robotic movement potentially approaches human thinking in more complex situations and focuses on the new human-robot interaction scheme by proposing a weak human preference supervision framework in deep RL for robotic controls. Our breakthrough results are part of a broader core set of robotics challenges concerning the deployment of automated driving systems.

\section*{Acknowledgements}
We thank Prof. Jun Wang (Department of Computer Science, University College London, UK) and Prof. Haifeng Zhang (Institute of Automation, Chinese Academy of Sciences, China) for providing suggestions to improve the presentation of the manuscript. This work was supported in part by the CoSE incentive grant scheme at the University of Tasmania and the US Office of Naval Research Global, under Cooperative Agreement Number: ONRG-NICOP-N62909-19-1-2058.

\section*{Appendix}

The code of this paper can be found at GitHub https://github.com/kaichiuwong/rlhps 



\ifCLASSOPTIONcaptionsoff
  \newpage
\fi



%

\bibliographystyle{IEEEtran}
\bibliography{IEEEabrv,bib}

\begin{thebibliography}{10}
\providecommand{\url}[1]{#1}
\csname url@samestyle\endcsname
\providecommand{\newblock}{\relax}
\providecommand{\bibinfo}[2]{#2}
\providecommand{\BIBentrySTDinterwordspacing}{\spaceskip=0pt\relax}
\providecommand{\BIBentryALTinterwordstretchfactor}{4}
\providecommand{\BIBentryALTinterwordspacing}{\spaceskip=\fontdimen2\font plus
\BIBentryALTinterwordstretchfactor\fontdimen3\font minus
  \fontdimen4\font\relax}
\providecommand{\BIBforeignlanguage}[2]{{%
\expandafter\ifx\csname l@#1\endcsname\relax
\typeout{** WARNING: IEEEtran.bst: No hyphenation pattern has been}%
\typeout{** loaded for the language `#1'. Using the pattern for}%
\typeout{** the default language instead.}%
\else
\language=\csname l@#1\endcsname
\fi
#2}}
\providecommand{\BIBdecl}{\relax}
\BIBdecl

\bibitem{mnih2015human}
V.~Mnih, K.~Kavukcuoglu, D.~Silver, A.~A. Rusu, J.~Veness, M.~G. Bellemare,
  A.~Graves, M.~Riedmiller, A.~K. Fidjeland, G.~Ostrovski \emph{et~al.},
  ``Human-level control through deep reinforcement learning,'' \emph{nature},
  vol. 518, no. 7540, pp. 529--533, 2015.

\bibitem{nikolaidis2017game}
S.~Nikolaidis, S.~Nath, A.~D. Procaccia, and S.~Srinivasa, ``Game-theoretic
  modeling of human adaptation in human-robot collaboration,'' in
  \emph{Proceedings of the 2017 ACM/IEEE international conference on
  human-robot interaction}, 2017, pp. 323--331.

\bibitem{bogert2016expectation}
K.~Bogert, J.~F.-S. Lin, P.~Doshi, and D.~Kulic, ``Expectation-maximization for
  inverse reinforcement learning with hidden data,'' in \emph{Proceedings of
  the 2016 International Conference on Autonomous Agents \& Multiagent
  Systems}, 2016, pp. 1034--1042.

\bibitem{lu2020bearing}
Y.~Lu, C.~Wen, T.~Shen, and W.~Zhang, ``Bearing-based adaptive neural formation
  scaling control for autonomous surface vehicles with uncertainties and input
  saturation,'' \emph{IEEE Transactions on Neural Networks and Learning
  Systems}, 2020.

\bibitem{xu2019design}
S.~Xu and H.~Peng, ``Design, analysis, and experiments of preview path tracking
  control for autonomous vehicles,'' \emph{IEEE Transactions on Intelligent
  Transportation Systems}, vol.~21, no.~1, pp. 48--58, 2019.

\bibitem{russell2016should}
S.~Russell, ``Should we fear supersmart robots?'' \emph{Scientific American},
  vol. 314, no.~6, pp. 58--59, 2016.

\bibitem{amodei2016concrete}
D.~Amodei, C.~Olah, J.~Steinhardt, P.~Christiano, J.~Schulman, and D.~Man{\'e},
  ``Concrete problems in ai safety,'' \emph{arXiv preprint arXiv:1606.06565},
  2016.

\bibitem{ke2020enhancing}
Z.~Ke, Z.~Li, Z.~Cao, and P.~Liu, ``Enhancing transferability of deep
  reinforcement learning-based variable speed limit$\backslash$endgraf control
  using transfer learning,'' \emph{IEEE Transactions on Intelligent
  Transportation Systems}, 2020.

\bibitem{cao2019reinforcement}
Z.~Cao and C.-T. Lin, ``Reinforcement learning from hierarchical critics,''
  \emph{arXiv preprint arXiv:1902.03079}, 2019.

\bibitem{cao2020hierarchical}
Z.~Cao, K.~Wong, Q.~Bai, and C.-T. Lin, ``Hierarchical and non-hierarchical
  multi-agent interactions based on unity reinforcement learning,'' in
  \emph{Proceedings of the 19th International Conference on Autonomous Agents
  and MultiAgent Systems}, 2020, pp. 2095--2097.

\bibitem{RN55}
A.~Y. Ng and S.~J. Russell, ``Algorithms for inverse reinforcement learning,''
  in \emph{Icml}, vol.~1, Conference Proceedings, pp. 663--670.

\bibitem{RN59}
A.~Hussein, M.~M. Gaber, E.~Elyan, and C.~Jayne, ``Imitation learning: A survey
  of learning methods,'' \emph{ACM Computing Surveys (CSUR)}, vol.~50, no.~2,
  pp. 1--35, 2017.

\bibitem{schroecker2017state}
Y.~Schroecker and C.~L. Isbell, ``State aware imitation learning,'' in
  \emph{Advances in Neural Information Processing Systems}, 2017, pp.
  2911--2920.

\bibitem{RN52}
P.~F. Christiano, J.~Leike, T.~Brown, M.~Martic, S.~Legg, and D.~Amodei, ``Deep
  reinforcement learning from human preferences,'' in \emph{Advances in Neural
  Information Processing Systems}, 2017, Conference Proceedings, pp.
  4299--4307.

\bibitem{xiao2020novel}
F.~Xiao, Z.~Cao, and A.~Jolfaei, ``A novel conflict measurement in decision
  making and its application in fault diagnosis,'' \emph{IEEE Transactions on
  Fuzzy Systems}, 2020.

\bibitem{RN61}
E.~Todorov, T.~Erez, and Y.~Tassa, ``Mujoco: A physics engine for model-based
  control,'' in \emph{2012 IEEE/RSJ International Conference on Intelligent
  Robots and Systems}.\hskip 1em plus 0.5em minus 0.4em\relax IEEE, Conference
  Proceedings, pp. 5026--5033.

\bibitem{woodworth2018preference}
B.~Woodworth, F.~Ferrari, T.~E. Zosa, and L.~D. Riek, ``Preference learning in
  assistive robotics: Observational repeated inverse reinforcement learning,''
  in \emph{Machine Learning for Healthcare Conference}, 2018, pp. 420--439.

\bibitem{furnkranz2012preference}
J.~F{\"u}rnkranz, E.~H{\"u}llermeier, W.~Cheng, and S.-H. Park,
  ``Preference-based reinforcement learning: a formal framework and a policy
  iteration algorithm,'' \emph{Machine learning}, vol.~89, no. 1-2, pp.
  123--156, 2012.

\bibitem{aggarwal2019modelling}
M.~Aggarwal and A.~Fallah~Tehrani, ``Modelling human decision behaviour with
  preference learning,'' \emph{INFORMS Journal on Computing}, vol.~31, no.~2,
  pp. 318--334, 2019.

\bibitem{palan2019learning}
M.~Palan, N.~C. Landolfi, G.~Shevchuk, and D.~Sadigh, ``Learning reward
  functions by integrating human demonstrations and preferences,'' \emph{arXiv
  preprint arXiv:1906.08928}, 2019.

\bibitem{wirth2017survey}
C.~Wirth, R.~Akrour, G.~Neumann, and J.~F{\"u}rnkranz, ``A survey of
  preference-based reinforcement learning methods,'' \emph{The Journal of
  Machine Learning Research}, vol.~18, no.~1, pp. 4945--4990, 2017.

\bibitem{kreps1988notes}
D.~Kreps, \emph{Notes on the Theory of Choice}.\hskip 1em plus 0.5em minus
  0.4em\relax Westview press, 1988.

\bibitem{wilson2012bayesian}
A.~Wilson, A.~Fern, and P.~Tadepalli, ``A bayesian approach for policy learning
  from trajectory preference queries,'' in \emph{Advances in neural information
  processing systems}, 2012, pp. 1133--1141.

\bibitem{ho2016model}
J.~Ho, J.~Gupta, and S.~Ermon, ``Model-free imitation learning with policy
  optimization,'' in \emph{International Conference on Machine Learning}, 2016,
  pp. 2760--2769.

\bibitem{schulman2017proximal}
J.~Schulman, F.~Wolski, P.~Dhariwal, A.~Radford, and O.~Klimov, ``Proximal
  policy optimization algorithms,'' \emph{arXiv preprint arXiv:1707.06347},
  2017.

\bibitem{todorov2012MuJoCo}
E.~Todorov, T.~Erez, and Y.~Tassa, ``Mujoco: A physics engine for model-based
  control,'' in \emph{2012 IEEE/RSJ International Conference on Intelligent
  Robots and Systems}.\hskip 1em plus 0.5em minus 0.4em\relax IEEE, 2012, pp.
  5026--5033.

\bibitem{abadi2016tensorflow}
M.~Abadi, P.~Barham, J.~Chen, Z.~Chen, A.~Davis, J.~Dean, M.~Devin,
  S.~Ghemawat, G.~Irving, M.~Isard \emph{et~al.}, ``Tensorflow: A system for
  large-scale machine learning,'' in \emph{12th $\{$USENIX$\}$ symposium on
  operating systems design and implementation ($\{$OSDI$\}$ 16)}, 2016, pp.
  265--283.

\bibitem{brockman2016openai}
G.~Brockman, V.~Cheung, L.~Pettersson, J.~Schneider, J.~Schulman, J.~Tang, and
  W.~Zaremba, ``Openai gym,'' \emph{arXiv preprint arXiv:1606.01540}, 2016.

\bibitem{RN64}
Y.~Tassa, Y.~Doron, A.~Muldal, T.~Erez, Y.~Li, D.~d.~L. Casas, D.~Budden,
  A.~Abdolmaleki, J.~Merel, and A.~Lefrancq, ``Deepmind control suite,''
  \emph{arXiv preprint arXiv:1801.00690}, 2018.

\end{thebibliography}

%







\end{document}